\documentclass[10pt,twocolumn,letterpaper]{article}

\usepackage{cvpr}
\usepackage{times}
\usepackage{epsfig}
\usepackage{graphicx}
\usepackage{amsmath}
\usepackage{amssymb}
\usepackage{enumitem}
\usepackage{wrapfig}
\usepackage{tablefootnote}
\usepackage{makecell}
\usepackage{booktabs}
\usepackage{adjustbox}

\newcommand{\ra}[1]{\renewcommand{\arraystretch}{#1}}

\usepackage[pagebackref=true,breaklinks=true,letterpaper=true,colorlinks,bookmarks=false]{hyperref}

\cvprfinalcopy % *** Uncomment this line for the final submission

\def\cvprPaperID{****} % *** Enter the CVPR Paper ID here

\ifcvprfinal\fi
\begin{document}

%%%%%%%%% TITLE
\title{BlazePose: On-device Real-time Body Pose tracking}

\author{
Valentin Bazarevsky\qquad Ivan Grishchenko\qquad Karthik Raveendran\\Tyler Zhu\qquad Fan Zhang\qquad  Matthias Grundmann\\
Google Research\\
1600 Amphitheatre Pkwy, Mountain View, CA 94043, USA\\
{\tt\small \{valik, igrishchenko, krav, tylerzhu, zhafang, grundman\}@google.com}\\
}

\maketitle
%\thispagestyle{empty}

%%%%%%%%% ABSTRACT
\begin{abstract}

We present BlazePose, a lightweight convolutional neural network architecture for human pose estimation that is tailored for real-time inference on mobile devices. During inference, the network produces 33 body keypoints for a single person and runs at over 30 frames per second on a Pixel 2 phone. This makes it particularly suited to real-time use cases like fitness tracking and sign language recognition. Our main contributions include a novel body pose tracking solution and a lightweight body pose estimation neural network that uses both heatmaps and regression to keypoint coordinates.
\end{abstract}

%%%%%%%%% BODY TEXT
\section{Introduction} \label{sect:introduction}

Human body pose estimation from images or video plays a central role in various applications such as health tracking, sign language recognition, and gestural control. This task is challenging due to a wide variety of poses, numerous degrees of freedom, and occlusions. Recent work \cite{DeepHighResolutionPosePaper}\cite{PifPafPaper} has shown significant progress on pose estimation. The common approach is to produce heatmaps for each joint along with refining offsets for each coordinate. While this choice of heatmaps scales to multiple people with minimal overhead, it makes the model for a single person considerably larger than is suitable for real-time inference on mobile phones. In this paper, we address this particular use case and demonstrate significant speedup of the model with little to no quality degradation.

In contrast to heatmap-based techniques, regression-based approaches, while less computationally demanding and more scalable, attempt to predict the mean coordinate values, often failing to address the underlying ambiguity. Newell \etal \cite{stacked_hourglass} have shown that the stacked hourglass architecture gives a significant boost to the quality of the prediction, even with a smaller number of parameters. We extend this idea in our work and use an encoder-decoder network architecture to predict heatmaps for all joints, followed by another encoder that regresses directly to the coordinates of all joints. The key insight behind our work is that the heatmap branch can be discarded during inference, making it sufficiently lightweight to run on a mobile phone.
%-------------------------------------------------------------------------
\section{Model Architecture and Pipeline Design} 
\label{sect:architecture}

%-------------------------------------------------------------------------
\subsection{Inference pipeline}
\label{subsect:inference}
\begin{figure}
  \centering
  \includegraphics[width=\linewidth]{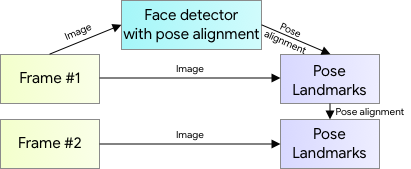}
  \caption{\protect\centering Inference pipeline. See text.}
  \label{fig:blazepose_pipeline}
\end{figure}

During inference, we employ a detector-tracker setup (see Figure \ref{fig:blazepose_pipeline}), which shows excellent real-time performance on a variety of tasks such as hand landmark prediction \cite{MediaPipeHandsAIBlog} and dense face landmark prediction \cite{facemesh}. Our pipeline consists of a lightweight body pose detector followed by a pose tracker network. The tracker predicts keypoint coordinates, the presence of the person on the current frame, and the refined region of interest for the current frame. When the tracker indicates that there is no human present, we re-run the detector network on the next frame.

\subsection{Person detector}

The majority of modern object detection solutions rely on the Non-Maximum Suppression (NMS) algorithm for their last post-processing step. This works well for rigid objects with few degrees of freedom. However, this algorithm breaks down for scenarios that include highly articulated poses like those of humans, \eg people waving or hugging. This is because multiple, ambiguous boxes satisfy the intersection over union (IoU) threshold for the NMS algorithm. To overcome this limitation, we focus on detecting the bounding box of a relatively rigid body part like the human face or torso. We observed that in many cases, the strongest signal to the neural network about the position of the torso is the person's face (as it has high-contrast features and has fewer variations in appearance). To make such a person detector fast and lightweight, we make the strong, yet for AR applications valid, assumption that the head of the person should always be visible for our single-person use case.
\begin{wrapfigure}{r}{0pt}
  \centering
  \includegraphics[width=0.4\linewidth]{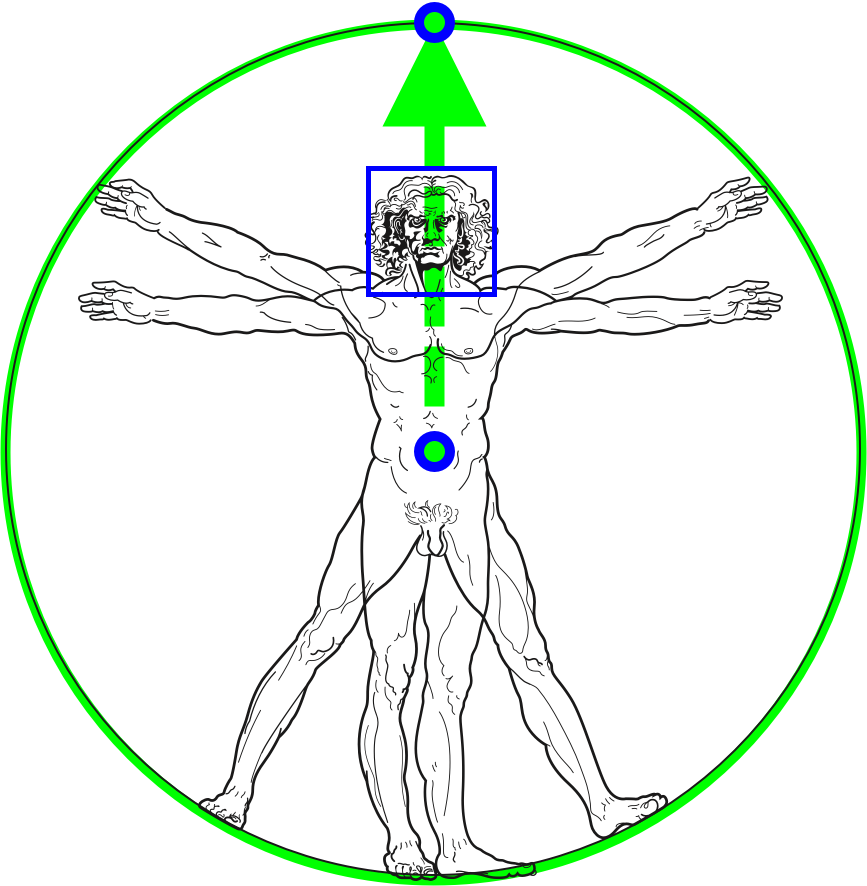}
  \caption{\protect\centering Vitruvian man aligned via our detector vs. face detection bounding box. See text for details.}
  \label{fig:virtuvian}
\end{wrapfigure}
As a consequence, we use a fast on-device face detector \cite{BlazeFace} as a proxy for a person detector. This face detector predicts additional person-specific alignment parameters: the middle point between the person's hips, the size of the circle circumscribing the whole person, and incline (the angle between the lines connecting the two mid-shoulder and mid-hip points).

%-------------------------------------------------------------------------
\subsection{Topology}

We present a new topology using 33 points on the human body by taking the superset of those used by BlazeFace\cite{BlazeFace}, BlazePalm\cite{MediaPipeHandsAIBlog}, and Coco\cite{CocoPaper}. This allows us to be consistent with the respective datasets and inference networks.
\begin{figure}
  \centering
  \includegraphics[width=0.5\linewidth]{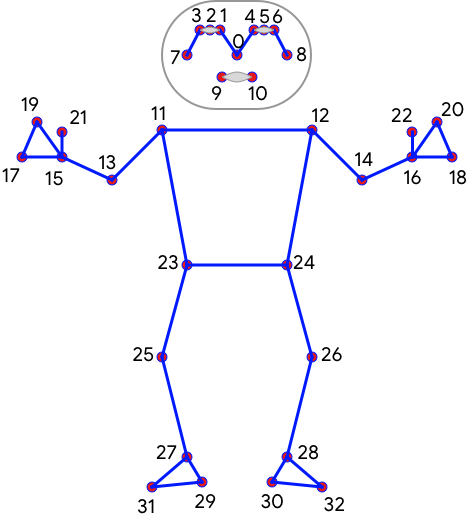}
  \caption{\protect\centering 33 keypoint topology.}
  \label{fig:bb_33topology}
\end{figure}

In contrast with the OpenPose\cite{OpenPosePaper} and Kinect\cite{Kinect} topologies, we use only a minimally sufficient number of keypoints on the face, hands, and feet to estimate rotation, size, and position of the region of interest for the subsequent model. The topology we use is shown in Figure \ref{fig:bb_33topology}. For additional information, please see Appendix A.

\subsection{Dataset}
Compared to the majority of existing pose estimation solutions that detect keypoints using heatmaps, our tracking-based solution requires an initial pose alignment. We restrict our dataset to those cases where either the whole person is visible, or where hips and shoulders keypoints can be confidently annotated. To ensure the model supports heavy occlusions that are not present in the dataset, we use substantial occlusion-simulating augmentation. Our training dataset consists of 60K images with a single or few people in the scene in common poses and 25K images with a single person in the scene performing fitness exercises. All of these images were annotated by humans.

%-------------------------------------------------------------------------
\subsection{Neural network architecture}
The pose estimation component of our system predicts the location of all 33 person keypoints, and uses the person alignment proposal provided by the first stage of the pipeline (Section \ref{subsect:inference}). 

\begin{figure}
  \centering
  \includegraphics[width=0.8\linewidth]{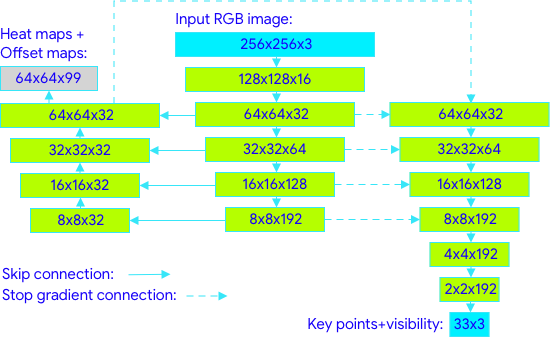}
  \caption{Network architecture. See text for details.}
  \label{fig:tracker_net}
\end{figure}

We adopt a combined heatmap, offset, and regression approach, as shown in Figure \ref{fig:tracker_net}. We use the heatmap and offset loss only in the training stage and remove the corresponding output layers from the model before running the inference. Thus, we effectively use the heatmap to supervise the lightweight embedding, which is then utilized by the regression encoder network. This approach is partially inspired by Stacked Hourglass approach of Newell et al. \cite{stacked_hourglass}, but in our case, we stack a tiny encoder-decoder heatmap-based network and a subsequent regression encoder network.

We actively utilize skip-connections between all the stages of the network to achieve a balance between high- and low-level features. However, the gradients from the regression encoder are not propagated back to the heatmap-trained features (note the gradient-stopping connections in Figure \ref{fig:tracker_net}). We have found this to not only improve the heatmap predictions, but also substantially increase the coordinate regression accuracy.

\subsection{Alignment and occlusions augmentation}

A relevant pose prior is a vital part of the proposed solution. We deliberately limit supported ranges for the angle, scale, and translation during augmentation and data preparation when training. This allows us to lower the network capacity, making the network faster while requiring fewer computational and thus energy resources on the host device.

Based on either the detection stage or the previous frame keypoints, we align the person so that the point between the hips is located at the center of the square image passed as the neural network input. We estimate rotation as the line $L$ between mid-hip and mid-shoulder points and rotate the image so $L$ is parallel to the y-axis. The scale is estimated so that all the body points fit in a square bounding box circumscribed around the body, as shown in Figure \ref{fig:virtuvian}. On top of that, we apply 10\% scale and shift augmentations to ensure the tracker handles body movements between the frames and distorted alignment. 

\begin{wrapfigure}{r}{0pt}
  \centering
  \includegraphics[height=0.5\linewidth]{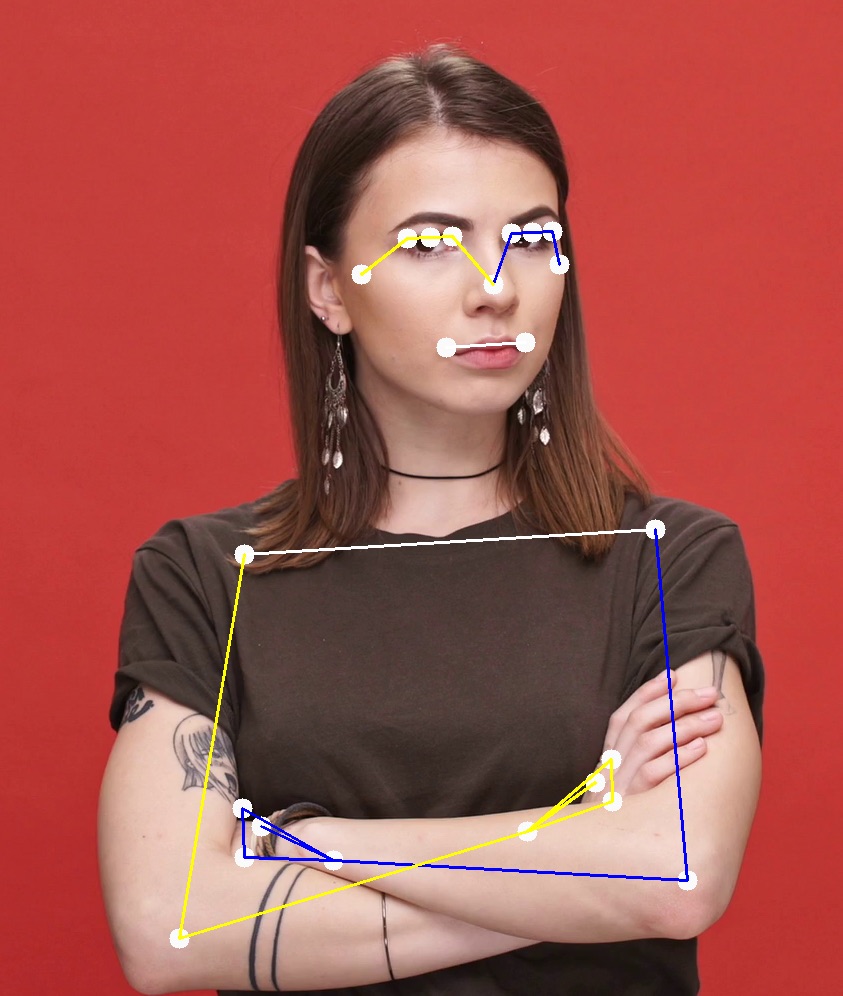}
  \caption{\protect\centering BlazePose results on upper-body case}
  \label{fig:upperbody}
\end{wrapfigure}

To support the prediction of invisible points, we simulate occlusions (random rectangles filled with various colors) during training and introduce a per-point visibility classifier that indicates whether a particular point is occluded and if the position prediction is deemed inaccurate. This allows tracking a person constantly even for cases of significant occlusions, like upper body-only or when the majority of person body is out of scene as shown on Figure \ref{fig:upperbody}.

%-------------------------------------------------------------------------
\section{Experiments} \label{sect:experiments}

To evaluate our model's quality, we chose OpenPose \cite{OpenPosePaper} as a baseline. To that end, we manually annotated two in-house datasets of 1000 images, each with 1--2 people in the scene.  The first dataset, referred to as AR dataset, consist of a wide variety of human poses in the wild, while the second is comprised of yoga/fitness poses only. For consistency, we only used MS Coco \cite{CocoPaper} topology with 17 points for evaluation, which is a common subset of both OpenPose and BlazePose. As an evaluation metric, we use the Percent of Correct Points with 20\% tolerance (PCK@0.2) (where we assume the point to be detected correctly if the 2D Euclidean error is smaller than 20\% of the corresponding person's torso size). To verify the human baseline, we asked two annotators to re-annotate the AR dataset independently and obtained an average PCK@0.2 of 97.2.

\begin{table}[ht]
\begin{center}
\ra{1.2}
\begin{tabular}{llll}
\toprule
\thead{Model} & \thead{FPS} & \thead{AR Dataset, \\ PCK@0.2} & \thead{Yoga Dataset, \\ PCK@0.2} \\
\midrule
OpenPose (body only) & 0.4\tablefootnote{Desktop CPU with 20 cores (Intel i9-7900X)} & \textbf{87.8} & 83.4 \\
BlazePose Full & 10\tablefootnote{Pixel 2 Single Core via XNNPACK backend\label{XNNPACK_EVAL}} & 84.1 & \textbf{84.5} \\
BlazePose Lite & \textbf{31}\textsuperscript{\getrefnumber{XNNPACK_EVAL}} & 79.6 & 77.6 \\
\bottomrule
\end{tabular}
\end{center}
\caption{BlazePose vs OpenPose}
\label{tbl:more_speed}
\end{table}

We trained two models with different capacities: BlazePose Full (6.9 MFlop, 3.5M Params) and BlazePose Lite (2.7 MFlop, 1.3M Params). Although our models show slightly worse performance than the OpenPose model on the AR dataset, BlazePose Full outperforms OpenPose on Yoga/Fitness use cases. At the same time, BlazePose performs 25--75 times faster on a \emph{single mid-tier phone CPU} compared to OpenPose on a \emph{20 core desktop} CPU\cite{OpenPoseBenchmark} depending on the requested quality. 
%-------------------------------------------------------------------------

%-------------------------------------------------------------------------
\section{Applications} \label{sect:applications}

\begin{figure}
  \centering
  \includegraphics[width=0.49\linewidth, height=0.49\linewidth]{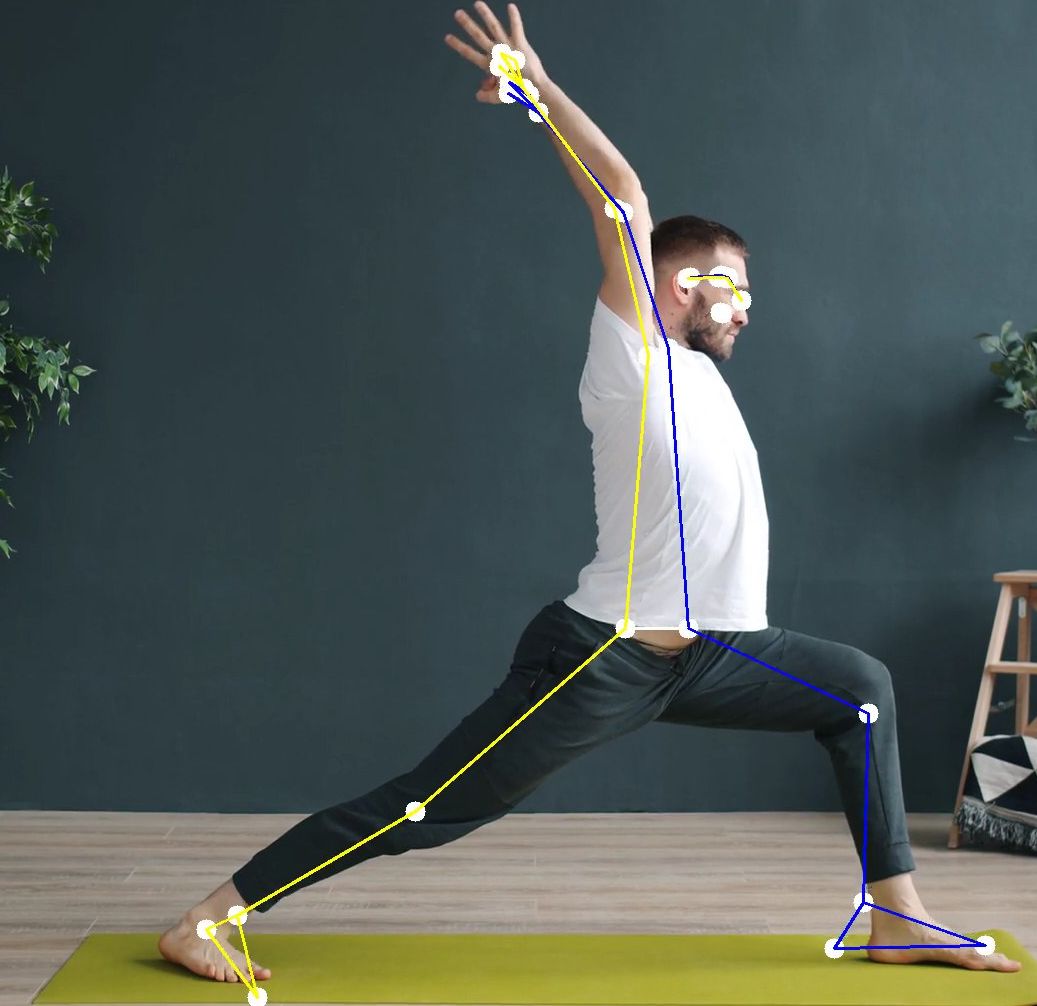}
  \includegraphics[width=0.49\linewidth, height=0.49\linewidth]{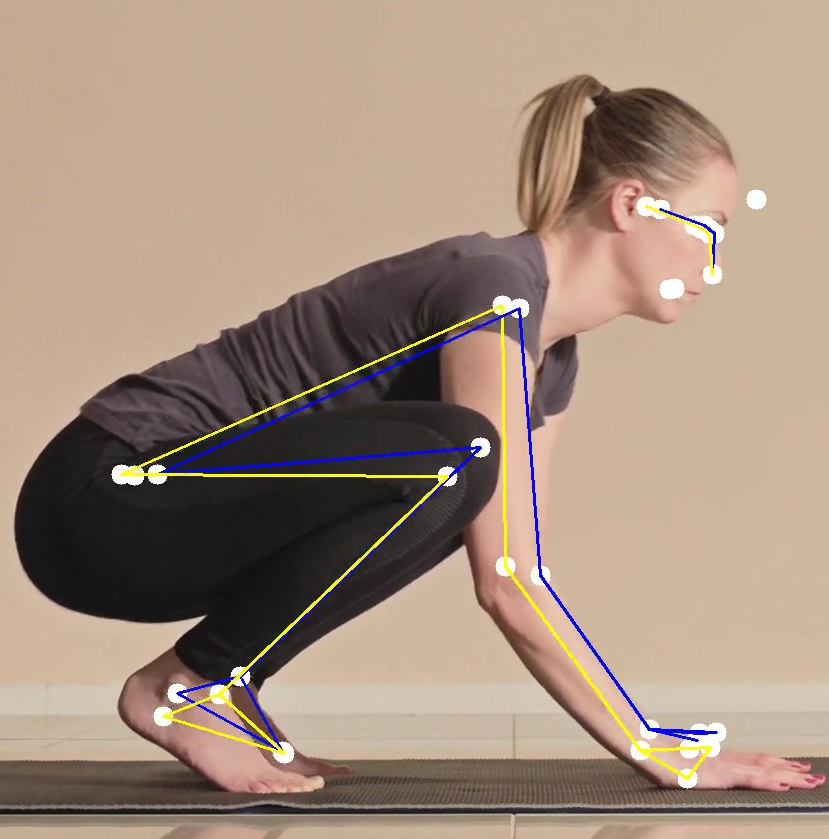}
  \includegraphics[width=0.49\linewidth, height=0.49\linewidth]{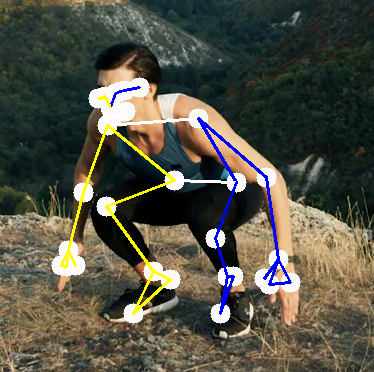}
  \includegraphics[width=0.49\linewidth, height=0.49\linewidth]{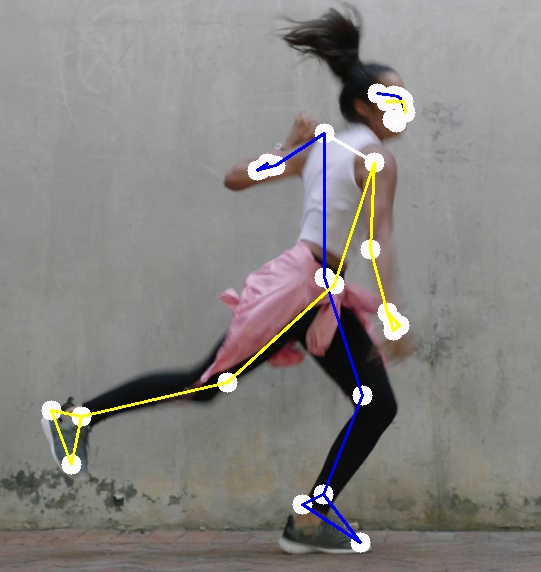}
  \caption{\protect\centering BlazePose results on yoga and fitness poses.}
  \label{fig:yoga1}
\end{figure}

We developed this new, on-device, single person-specific human pose estimation model to enable various performance-demanding use cases such as Sign Language, Yoga/Fitness tracking and AR. This model works in near-realtime on a mobile CPU and can be sped up to super-realtime latency on a mobile GPU. As its 33 keypoint topology is consistent with BlazeFace\cite{BlazeFace} and BlazePalm\cite{MediaPipeHandsAIBlog}, it can be a backbone for subsequent hand pose\cite{MediaPipeHandsAIBlog} and facial geometry estimation\cite{facemesh} models.

Our approach natively scales to a bigger number of keypoints, 3D support, and additional keypoint attributes, since it is not based on heatmaps/offset maps and therefore does not require an additional full-resolution layer per each new feature type.

{\small
\bibliographystyle{ieee_fullname}
\bibliography{egbib}
}

\newpage
\section*{Appendix A. BlazePose keypoint names} \label{sect:appendixa}

\vspace{\topsep}
\begin{enumerate}
\setlength{\parskip}{0pt}
\setlength{\itemsep}{0pt plus 3pt}
  \setcounter{enumi}{-1}
  \item Nose
  \item Left eye inner
  \item Left eye
  \item Left eye outer
  \item Right eye inner
  \item Right eye
  \item Right eye outer
  \item Left ear
  \item Right ear
  \item Mouth left
  \item Mouth right
  \item Left shoulder
  \item Right shoulder
  \item Left elbow
  \item Right elbow
  \item Left wrist
  \item Right wrist
  \item Left pinky \#1 knuckle
  \item Right pinky \#1 knuckle
  \item Left index \#1 knuckle
  \item Right index \#1 knuckle
  \item Left thumb \#2 knuckle
  \item Right thumb \#2 knuckle
  \item Left hip
  \item Right hip
  \item Left knee
  \item Right knee
  \item Left ankle
  \item Right ankle
  \item Left heel
  \item Right heel
  \item Left foot index
  \item Right foot index
\end{enumerate}
\vspace{-\topsep}

\end{document}